\documentclass{article} % DO NOT CHANGE THIS
\usepackage{aaai2026}  % DO NOT CHANGE THIS
\usepackage{times}  % DO NOT CHANGE THIS
\usepackage{helvet}  % DO NOT CHANGE THIS
\usepackage{courier}  % DO NOT CHANGE THIS
\usepackage[hyphens]{url}  % DO NOT CHANGE THIS
\usepackage{graphicx} % DO NOT CHANGE THIS
\usepackage{natbib}  % DO NOT CHANGE THIS AND DO NOT ADD ANY OPTIONS TO IT
\usepackage{caption} % DO NOT CHANGE THIS AND DO NOT ADD ANY OPTIONS TO IT
\usepackage{amssymb}
\usepackage{algorithm}
\usepackage{algpseudocode}
\usepackage{amsmath}
\usepackage{subcaption}
\usepackage{bbold}
\usepackage{cancel}

\DeclareMathOperator*{\argmax}{arg\,max}

\newcommand{\cP}{\mathcal{P}}
\newcommand{\cO}{\mathcal{O}}
\newcommand{\cF}{\mathcal{F}}
\newcommand{\NLM}{\mathrm{NLM}}
\usepackage{newfloat}
\usepackage{listings}
\DeclareCaptionStyle{ruled}{labelfont=normalfont,labelsep=colon,strut=off} % DO NOT CHANGE THIS
\floatstyle{ruled}
\newfloat{listing}{tb}{lst}{}
\floatname{listing}{Listing}
\title{Neuro-Logic Lifelong Learning}
\author {
    Bowen He\textsuperscript{$*$},
    Xiaoan Xu,
    Alper Kamil Bozkurt,
    Vahid Tarokh,
    Juncheng Dong\textsuperscript{$\dagger$}
}
\affiliations {
    {
    Duke University\\[4pt]
    $*$\,\texttt{Corresponding author:bowen.he@duke.edu}\\
    $\dagger$\,\texttt{Advising}
    }
}

\usepackage{bibentry}
\begin{document}

\maketitle

\begin{abstract}
Solving Inductive Logic Programming (ILP) problems with neural networks is a key challenge in Neural-Symbolic Artificial Intelligence (AI). While most research has focused on designing novel network architectures for individual problems, less effort has been devoted to exploring new learning paradigms involving a sequence of problems. In this work, we investigate lifelong learning ILP, which leverages the compositional and transferable nature of logic rules for efficient learning of new problems. We introduce a compositional framework, demonstrating how logic rules acquired from earlier tasks can be efficiently reused in subsequent ones, leading to improved scalability and performance. We formalize our approach and empirically evaluate it on sequences of tasks. Experimental results validate the feasibility and advantages of this paradigm, opening new directions for continual learning in Neural-Symbolic AI.

\end{abstract}


\section{Introduction}
Neuro-Symbolic Artificial Intelligence~\citep{santoro2017simple,manhaeve2018deepproblogneuralprobabilisticlogic, dai2019bridging, garcez2020neurosymbolicai3rdwave, amizadeh2020neuro} has emerged as a promising research direction that combines modern neural networks with classic symbolic methods, thereby leveraging the strengths of both. At a high level, neural networks offer the expressivity and end-to-end learning capabilities needed to tackle complex problems where traditional symbolic methods can fall short. Meanwhile, symbolic approaches contribute advantages such as explicit representation of knowledge and reasoning, in which neural networks often underperform~\citep{valmeekam2023planningabilitieslargelanguage, li2024llmsrelationalreasoningfar, sheth2023neurosymbolicaiwhy}. Although the term \textit{Neuro-Symbolic Artificial Intelligence} spans a wide range of problems, paradigms, and methodologies, this work focuses specifically on tasks within the field of \textbf{inductive logic programming} (ILP)~\citep{cropper2022inductivelogicprogramming30}. Unlike standard logic programming, which draws conclusions from a given set of rules, ILP seeks to learn first-order logic rules that best explain observed examples from given relevant background knowledge.

To solve ILP problems using neural networks, researchers have introduced a variety of methods and compared them against traditional ILP solvers~\citep{evans2018learningexplanatoryrulesnoisy, glanois2021neurosymbolichierarchicalruleinduction, payani2019inductivelogicprogrammingdifferentiable,dong2019neurallogicmachines, BADREDDINE2022103649, sen2022neuro, zimmer2023differentiablelogicmachines}, demonstrating that neural network-based methods are both robust to noisy data and more efficient for large-scale tasks. 

However, most existing works primarily focus on designing new model architectures, dedicating relatively little attention to investigate learning paradigms beyond the conventional individual tasks setting. To this end, this work takes the first step to investigate the transferability of knowledge between ILP problems. 
Our insight is that \textbf{\emph{logic rules, by their nature, are compositional and reusable}}. Specifically, a logic rule learned from one task can be naturally reused for another task within the same domain. Moreover, cognitive scientists have argued that humans learn, think, and reason in a symbolic manner, i.e.,\emph{``symbolic models of cognition were the dominant computational approaches of cognition"}~\citep{Castro2025.02.05.636732,Besold_Kühnberger_2023}. 
Thus, to achieve the remarkable capabilities of lifelong learning and meta-learning observed in human intelligence, we envision that the interplay of neural network and symbolic method presents a promising new direction for lifelong learning.

We instantiate the aforementioned insight by introducing a novel lifelong learning problem for ILP. 
We introduce a compositional structure for neural logic models and evaluate their performance across sequences of tasks. By leveraging rules acquired from previous tasks, the neural logic models achieve significantly improved learning efficiency on new tasks.
In comparison to the existing works that primarily focus on the perspective of model parameter optimization---such as regularization-based~\citep{kirkpatrick2017overcoming, zenke2017continual}, experience replay-based~\citep{rolnick2019experience, buzzega2020dark}, and architecture-based approaches~\citep{rusu2016progressive, von2020continual, li2019learn}---lifelong learning for logic rules requires identifying which rules are beneficial for reuse 
and efficiently constructing new rules based on those already acquired. We take the first step in demonstrating the feasibility of this direction, paving the way for future research. Our empirical results confirm the enhanced learning efficiency achieved through lifelong learning. Furthermore, by simply incorporating experience replay, the model effectively retains its performance across tasks. Additionally, in certain experiments, we observe a backward transfer effect, where training on later tasks further improves performance on earlier tasks.

{\bf Contribution Statement.} We summarize our contributions as follows:
\begin{itemize}
    \item We formally introduce lifelong learning in ILP, framing it as a sequential optimization problem.
    \item We propose a neuro-symbolic approach that leverages the compositionality of logic rules to enable knowledge transfer across tasks.
    \item We validate on challenging logic reasoning tasks how \textbf{\emph{logic rule transfer improves learning efficiency}} and the acquisition of a common knowledge base during sequential learning.
\end{itemize}

{\bf Manuscript Organization.} We frist briefly review related works in Section~\ref{Related Works}. We introduce the definition of lifelong ILP problems in Section~\ref{Formulation}. We elaborate on our implementation in Section~\ref{Implementation}. After presenting our experiment results in Section~\ref{Experiments}, we conclude with a discussion for future directions.

% {\bf Manuscript Organization.} We frist briefly review related works in Section~\ref{Related Works}. We introduce the definition of lifelong ILP problems in Section~\ref{Formulation}. We elaborate on our implementation in Section~\ref{Implementation}. After presenting our experiment results in Section~\ref{Experiments}, we conclude with a discussion for future directions.

\section{Related Works} 
\label{Related Works}
We provide review for both inductive logic programming, extending the discussion to more recent neural network based approcach, and lifelong learning that aims to build AI systems that accumulate knowledge in a seuquence of tasks.
\subsection{Inductive Logic Programming}
Inductive Logic Programming (ILP)~\citep{cropper2022inductivelogicprogramming30} is a longstanding and still unresolved challenge in artificial intelligence, characterized by its aim to solve problems through logical reasoning. Unlike statistical machine learning, where predications are based on statistical inference, ILP relies on logic inference and learns a logic program that could be used further to solve problems. Its integration with reinforcement learning (RL) further leads to the field of relational RL~\citep{dvzeroski2001relational}, where the policy comprises logical rules and decisions are made through logical inference. The goal of ILP is to design AI systems that not only solve problems but do so through logical reasoning, which recent literature shows is lacking in purely neural network-based approaches~\citep{valmeekam2023planningabilitieslargelanguage, li2024llmsrelationalreasoningfar}. This raises a critical question: can we develop neural-symbolic methods that leverage the scalability of neural networks while also solving tasks through sound logical reasoning?
\cite{evans2018learningexplanatoryrulesnoisy} made a pioneering step towards this objective by introducing $\partial$ILP, which integrates neural networks with inductive logic programming (ILP) and addresses 20 ILP tasks sourced from previous literature or designed by the authors themselves. They showed that, compared to traditional ILP methods, $\partial$ILP is more robust against mislabeled targets that typically impair the performance of conventional approaches, reflecting the generalization capabilities of neural networks. Furthermore, \cite{jiang2019neurallogicreinforcementlearning} extends $\partial$ILP to the reinforcement learning setting by incorporating logic predicates into the state and action spaces. They evaluate its performance on two tasks, Blocksworld and Gridworld, comparing it against MLP neural networks. Their results demonstrate that MLPs are prone to failure on these tasks represented with logic predicates.

Neural logic machine (NLM)~\citep{dong2019neurallogicmachines} represents another line of research that aims to design better neural network architectures for logic reasoning. They proposed a forward chaining approach to represent logical rules, effectively addressing the memory cost issues associated with handling a large number of objects, a challenge previously encountered by the $\partial$ILP method. Despite the fact that the logic rules learned by NLMs are not explicitly extractable for human readers, they continue to be the most widely used benchmark in further applications~\citep{wang2025imperativelearningselfsupervisedneurosymbolic} and related works. Differentiable Logic Machines~\citep{zimmer2023differentiablelogicmachines} builds upon NLM to replace MLPs used by NLM with soft logic operator, providing more interpretability in the cost more computational requirement.\\
\cite{campero2018logical} proposes to learning vector embeddings for both logic predicates and logic rules. It's further expended by \cite{glanois2021neurosymbolichierarchicalruleinduction} using the forward-chaining perspective as in NLM.

 Logic rules, by its nature, provide the property of compositionality, a new logic rule could be always built by composing rules that have been acquired before~\citep{lin2014bias}. Moreover, logic rules that are learned in a task could be naturally transferred to be used in another task from the same domain. This ability of knowledge transfer and lifelong learing has been considered as essential for human-like AI~\citep{lake2016buildingmachineslearnthink}. Building on this insight, we investigate the lifelong learning ability of neural logic programming.

\subsection{Lifelong Learning}
Lifelong learning or continual learning, has been proposed as a research direction for developing artificial intelligence systems capable of learning a sequence of tasks continuously~\citep{wang2024comprehensive}. Ideally, a lifelong learning agent should achieve both \textit{forward transfer}, where knowledge from earlier tasks benefits subsequent ones, and \textit{backward transfer}, where learning newer tasks enhances performance on previous ones. At the very least, it should mitigate catastrophic forgetting, a phenomenon where learning later tasks causes the model to lose knowledge acquired from earlier tasks.

Classic lifelong learning methods typically fall into four categories: \emph{(i) Regularization-based approaches}, which constrain parameter updates within a certain range to mitigate forgetting~\citep{kirkpatrick2017overcoming, zenke2017continual, li2017learning}); \emph{(ii) Replay-based approaches}, which store or generate data samples from past tasks and replay them during training on newer tasks~\citep{rebuffi2017icarl, shin2017continual, lopez2017gradient, chaudhry2018efficient}; \emph{(iii) Optimization-based methods}, which manipulate gradients to preserve previously acquired knowledge~\citep{zeng2019continual, farajtabar2020orthogonal, saha2021gradient}); and \emph{(iv) Architecture-based methods}, which expand or reconfigure model architectures to accommodate new tasks or transfer knowledge~\citep{rusu2016progressive, yoon2017lifelong}. In all cases, these methods are mainly rooted in neural networks and can be broadly categorized as strategies for optimizing model parameters or architectures. In contrast, approaching lifelong learning from a neuro-symbolic perspective offers additional opportunities by leveraging the properties of symbolic methods. Our work, therefore, falls within this emerging paradigm.

One work highly relevant to our study is \cite{mendez2022lifelongmachinelearningfunctionally}. They propose incorporating compositionality into lifelong training, enabling new tasks to benefit from previously learned neural modules. However, their formulation remains within the domain of pure neural networks and can be categorized as an architecture-based method for lifelong learning. In contrast, we emphasize that logic rules inherently provide compositionality, which can be leveraged for learning. Thus, our work serves as a strong instance of lifelong learning with compositionality, facilitating the systematic reuse and adaptation of learned knowledge across tasks. Another work that explores lifelong learning from a neuro-symbolic perspective is \cite{marconato2023neuro}. However, their focus is on extracting reusable concepts from sub-symbolic inputs while preventing reasoning shortcuts that could lead to incorrect symbolic knowledge. In contrast, our work centers on the transfer of logic rules, where learning and reusing these rules play a fundamental role.

\section{Neural Logic Lifelong Learning}\label{Formulation} 
\subsection{Problem Formulation}
We introduce our problem definition of ILP. 
We first define \emph{objects} and their corresponding \emph{types} within the domain of interest. Next, we define \emph{predicates} and \emph{operations}. Finally, we frame ILP as an optimization problem.
\subsubsection{Object Sets.} We denote the set of objects in a given domain as $\mathcal{O}$, while another set $\Lambda = \{\lambda_1, \lambda_2, ..., \lambda_n\}$ defines all possible types that these objects can take, namely, $\forall o \in \mathcal{O}, type(o) \in \Lambda$. A partition of $\mathcal{O}$ is thus induced by grouping objects of the same type into a subset. Formally, we partition the set of objects $\cO$ into subsets $\{\cO_
\lambda\}_{\lambda \in \Lambda}$, where
\begin{itemize}
    \item Each subset is nonempty: $\cO_\lambda\neq\varnothing$;
    \item Subsets are pairwise disjoint: $\cO_{\lambda} \cap \cO_{\lambda'} = \varnothing$ for all $\lambda \neq \lambda'$;
    \item Their union forms the entire set: $\cup_{\lambda\in \Lambda} \mathcal{O}_{\lambda} = \mathcal{O}$;
    \item Objects within a single subset share the same type: $\forall \lambda \in \Lambda, \forall o_i, o_j \in \mathcal{O}_{\lambda}, \ type(o_i) = type(o_j)$.
\end{itemize}

\subsubsection{Predicates.} Next we define predicates, which are the cores of ILP programs. Consider $N \in \mathbb{Z}_{\ge 0}$. A $N$-ary predicate $\cP$ is a binary-valued function $\cP: \cO(\cP)\rightarrow\{0,1\}$ where $\cO(\cP)$ is the Cartesian product of $N$ elements, each of which arbitrarily selected from $\{\mathcal{O}_{\lambda_1}, \mathcal{O}_{\lambda_2}, \dots, \mathcal{O}_{\lambda_n}\}$, that is, 
\[
\mathcal{O}(\cP) = \mathcal{O}_{\lambda_{i_1}} \times \mathcal{O}_{\lambda_{i_2}} \times \dots \times \mathcal{O}_{\lambda_{i_N}},
\]
where $\lambda_{i_k} \in \Lambda$ for all $1 \leq k \leq N$. 
Any set of $N$-ary predicates $\{\cP_1, \cP_2, \dots, \cP_m \}$ can be composed to form a new predicate $\widetilde{\cP}=F(\cP_1,\cP_2,\dots,\cP_m)$ with a \emph{logic rule} $F$. 

Consider an example with the object set $\mathcal{O} = \{o_1, o_2\}$ where $type(o_1)=type(o_2)$. We define two $1$-ary predicates $\cP_1$ and $\cP_2$ over $\cO$ as
$\mathcal{P}_1(o_1) = 0, \mathcal{P}_1(o_2) = 1$; $\mathcal{P}_2(o_1) = 1, \mathcal{P}_2(o_2) = 1.$
With $\cP_1$ and $\cP_2$ defined above, we can compose a new predicate $\cP_3$ defined as $\mathcal{P}_3(X) \leftarrow \mathcal{P}_1(X) \land \mathcal{P}_2(X)$. Here, the applied logic rule $F(\cP_1,\cP_2)$ is $\mathcal{P}_1(X) \land \mathcal{P}_2(X)$, leading to the valuation of $\mathcal{P}_3$ as
\[
    \mathcal{P}_3(o_1) = 0, \mathcal{P}_3(o_2) = 1.
\]
\subsubsection{ILP Problems.} Based on the definitions above, we are now ready to define ILPs. An ILP takes a set of \emph{background knowledge} $B$ as input. Here, $B$ is a set of $m$ predicates $B = \{\mathcal{P}_1, \mathcal{P}_2, ..., \mathcal{P}_m\}$. Given a known target predicate $\cP^\star$, the goal of an ILP program is to learn a logic rule $F^\star$ such that $F^\star(B)=\cP^\star$. We follow the conventions to assume complete knowledge of $B$, that is, each $\mathcal{P}_i$ from $B$ can be valuated on all possible inputs $o \in \cO(\cP_i)$, where $\cO(\cP_i)$ is the input space of $\cP_i$. 
We denote the space of all logic rules in consideration as $\cF$. An instance of the ILP problem is defined by a tuple $(\mathcal{O}, \Lambda, B, E)$, where $(\mathcal{O}$, $\Lambda$, $B)$ follow the previous definitions and $E=\{(o, y=\cP^\star(o))| o \in \mathcal{O}(\cP^\star), y\in \{0, 1\}\}$ describes the knowledge about the target predicate $\cP^\star$. 
The solution to the problem is a logic rule $\widehat{F}$ such that
\begin{equation}\label{eqn:obj-ind-ilp}
\begin{aligned}
&\widehat{F} \in \argmax_{F \in \cF} \sum_{o, y \in E}\mathbb{1}(\widehat{\cP}(o) = y) \quad \mathrm{s.t.}\quad \widehat{\cP} = F(B)
\end{aligned}
\end{equation}

\subsection{Lifelong Learning ILP}
Now we define the \emph{lifelong learning ILP problem} (L2ILP). Consider a sequence of target predicates $\{\cP_1^\star, \cP^\star_2\,\dots\}$ sharing the same background knowledge $B$. Intuitively, this means that we observe a common set of foundational knowledge from which different target conclusions are to be inferred. For example, in a medical diagnosis setting, the background knowledge $B$ could include general medical facts such as symptoms  and their possible causes. Each target predicate $\cP^\star_{i}$ could represent a different diagnostic task, such as fever, infection or other specific disease. We note that a naive approach to the above problem is to independently search for the optimal logic rule $\widehat{F}_t$ for each target predicate $\cP_t^\star$. Yet, this method would largely overlook the potentially shared structure of the predicates among target predicates and result in significant computational inefficiency. 
To this end, the goal of L2ILP is to efficiently find the logic rule $\widehat{F}_t$ for target predicate $\cP_t^\star$, using knowledge of the previously learned logic rules $\{\widehat{F}_1,\dots,\widehat{F}_{t-1}\}$ for composing the previous target predicates $\{\cP_1^\star,\dots,\cP^\star_{t-1}\}$. 
% Motivated by the common observation in computer vision that representations learned by deep learning models can be shared across tasks to facilitate learning. 
While there may exist multiple approaches for this purpose, \textbf{\emph{we propose to utilize the compositionality of logic rules and predicates by reusing the intermediate predicates composed during the learning of previous target predicates}}, thereby efficiently composing new predicates. 

\subsubsection{Motivating Example.} Consider an example of learning two target predicates on a graph reasoning task. A graph is described by a binary predicate $IsConnected(X, Y)$, additionally, several unary predicates are used to describe the colors of the nodes, such as $Red(X)$, $Yellow(X)$, $Blue(X)$, etc. A target predicate $AdjacentToRed(X)$ could be defined from a rule 
\[
AdjacentToRed(X) \leftarrow \exists Y IsConnected(X, Y) \land Red(Y)
\] 
Another target predicate $MultipleRed(X)$ could be similarly defined as
\[
\begin{array}{rl}
\textit{MultiRed}(X) \leftarrow \exists Y \exists Z\hspace{-1.0em}
& IsConnected(X, Y) \land Red(Y)) \land \\
& IsConnected(X, Z) \land Red(Z)) \land \\
& Is\_Not(Y, Z)
\end{array}
\]
Learning these two predicates separately is computationally wasteful because the two rules share a common structure.\\
Following this observation, we propose to solve the optimization problem through the reuse of logic rules across predicate functions, thus achieving forward transfer of knowledge required by lifelong learning.
\subsubsection{Problem Formulation.} Specifically, recall that $\cF$ is the set of all possible logic rules in considerable. The goal of L2ILP to find a \emph{shared knowledge base} $B_S \subset H(B)$ where $H(B) = \{F(B)|F \in \cF\}$ is the set of all possible predicates that can composed from the background knowledge $B$. At time step $t$ where the goal is to find a logic rule $\widehat{F}_t$ for $\cP_t^\star$, we can \emph{jointly} find $B_S$ and $\widehat{F}_t$ through the following optimization problem, 
\begin{equation}\label{eqn:obj-l2lip}
\begin{array}{rl}
\displaystyle \argmax_{B_S,\widehat{F}_t} 
& \displaystyle \sum_{t'=1}^t \sum_{o, y \in E_i} \mathbb{1}(\widehat{P}_{t'}(o) = y) \\[0.5em]
\text{s.t.}\;\; 
& \widehat{P}_{t'} = \widehat{F}_{t'}(B_S),\quad t' \in \{1,\dots,t-1\}; \\
& \widehat{P}_{t} = \widehat{F}_{t}(B_S),\quad \widehat{F}_t \in \widetilde{\cF}_t,
\end{array}
\end{equation}
where $\widehat{F}_{t'}$ for $\{1,\dots,t-1\}$ are the learned logic rules for the target predicates previous to $t$. 
% In particular, since $\widehat{F}_{t'}$ are fixed in Problem~\ref{eqn:obj-l2lip}, they serve as regularization for searching $B_S$. 
Problem~(\ref{eqn:obj-l2lip}) tries to identify a shared knowledge base $B_S$ that can be \textbf{\textit{(i)}} used by previously learned logical rules $\widehat{F}_{t'}$ for previous target predicts and \textbf{\textit{(ii)}} used to find the logical rule $\widehat{F}_t$ for the current target predicate $\cP_t^\star$. Notably, since L2ILP uses a shared knowledge base $B_S$ to employ the compositionality of logic rules for efficient learning, the search space for $\widehat{F}_t$ (i.e., $\widetilde{\cF}_t$) can be chosen to be a much smaller space than the space of all possible logic rules in consideration $\cF$, i.e, $|\widetilde{\cF}_t| \ll |\cF|$. \textbf{\textit{This can significantly increase the efficiency of learning}}. 

% Following the definition of the last section, we could naturally define the lifelong learning problem as a sequence of problem tuples
% $(P(\mathcal{O}, B), \mathcal{P}^\star_i)$ that generate example sets based on different target predicates, namely, the example set $E_i$ of a data point $(\mathcal{O}, \Lambda, B, E_i)$ is defined by the ith target predicate $\mathcal{P}^\star_i$. A solution to the problem will be a sequence of predicate functions $\mathcal{P}^{\star}_1,\mathcal{P}^{\star}_2,...,\mathcal{P}^{\star}_n$ such that,
% \begin{equation*}
% \begin{aligned}
% &\max_{\mathcal{P}^{\star}_1, \mathcal{P}^{\star}_2, ..., \mathcal{P}^{\star}_n, m_1, m_2, ..., m_n}\sum_{i=1}^n \mathbb{E}[\sum_{o, y \in E_i}\mathbf{1}(\mathcal{P}_i^{\star}(o) = y)],\\
% \text{subject to}\\
% &\mathcal{P}_1^{\star} \in H^{m_1}(B)\\
% &\mathcal{P}_2^{\star} \in H^{m_2}(B)\\
% &...\\
% &\mathcal{P}_n^{\star} \in H^{m_n}(B)
% \end{aligned}
% \end{equation*}

% Formally, a solution to a problem could be written as
\section{Compositional Neuro-Logic Model}\label{Implementation}
\subsubsection{Knowledge Base.} We implement L2ILP using Neural Logic Machine (NLM)~\citep{dong2019neurallogicmachines}. NLM chains together a sequence of logic layers (i.e., neural network layers with predicates as inputs and outputs), where each layer can be viewed as learning logical rules over the immediate input predicates. In our formulation, each NLM layer can be interpreted as a search space for logical rules, denoted as $\cF_{\NLM}$.
Rules in $\cF_{\NLM}$ are applied to the input predicates $B$ to construct the output predicate space, i.e., $H_{\NLM}(B)=\{F(B): F \in \cF_{\NLM}\}$.
Note that here the search space represented a NLM layer $\cF_{\NLM}$ is much smaller than the whole search space $\cF$. 
While one layer of NLM only generates target predicates with limited variation, chaining multiple layers lead to a complex and expressive target predicate space. 
% However, constraining $B_S$ to depend solely on $H(B)$ significantly limits the expressiveness of the knowledge base, as it only includes predicates that could be directly composed from the background knowledge. To address this issue, we construct the knowledge base to include NLM layers of varying depth. 
Specifically, the target predicate space with $n$ NLM layers can be defined recursively as 
\[
\begin{array}{rl}
H_{\NLM}^n(B) = \hspace{-0.9em}
& \{F(B \cup H_{\NLM}^{n-1}(B)) \mid F \in \cF\}, \\
\text{where}\hspace{-0.9em}\quad 
& H_{\NLM}^1(B) = H_{\NLM}(B)
\end{array}
\]
to represent the predicate space induced by iteratively applying logic rules from $\cF_{\NLM}$. 

\begin{figure}[htbp]
    \centering
    \hspace*{-0.5cm}
    \includegraphics[width=0.5\textwidth]{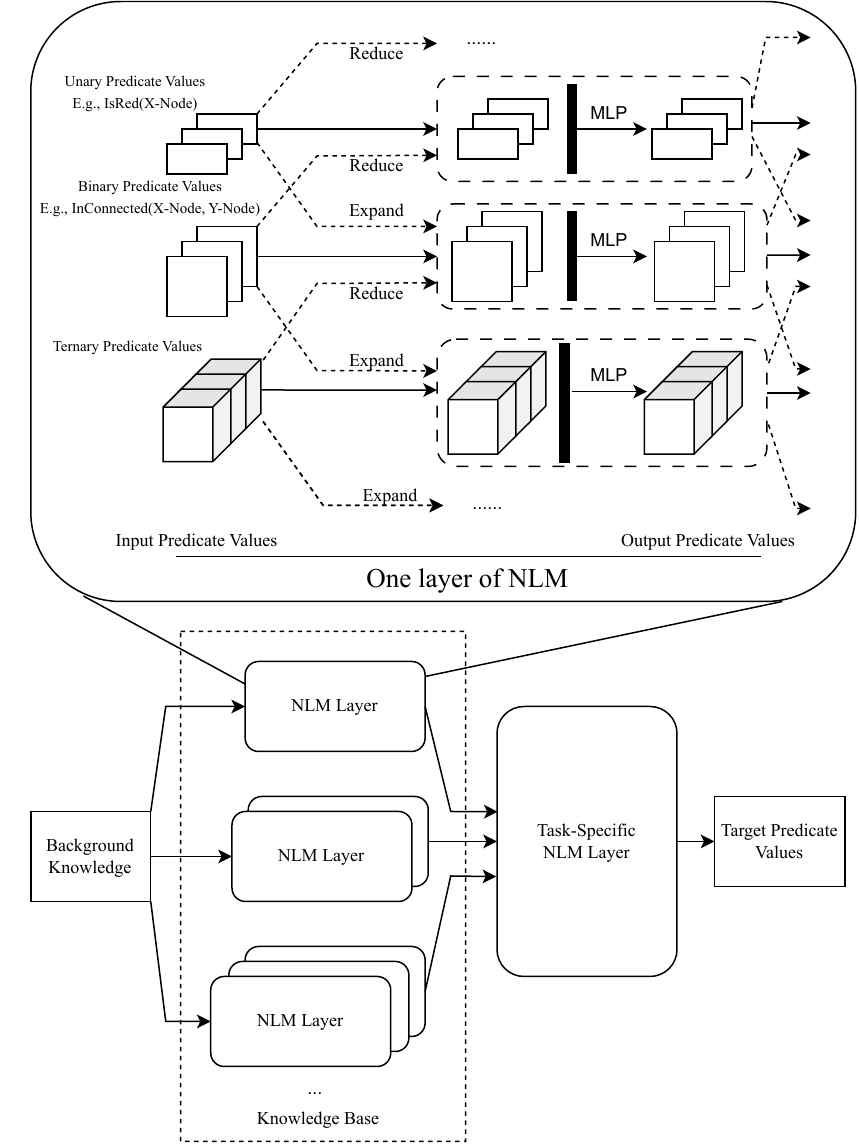}
    \caption{An illustration of Compositional Logic Model. Compositional Logic Model takes
object properties and relations as input and outputs relations of the objects.}
    \label{fig:NLMArch}
\end{figure}

We choose to construct the shared knowledge base $B_S \subset \bigcup_{i=1}^{n} H_{\NLM}^i(B)$ with the union of target predicates of increasing complexity $i \in \{1,\dots,n\}$ where $n$ is a hyperparameter to control the trade-off between the model complexity and target predicate expressiveness. In particular, by constructing the knowledge base with predicates of varying complexity, we achieve fine-grained predicate composition, reminiscent of the success of multi-scale representation-learning methods in the field of computer vision~\citep{fan2021multiscalevisiontransformers}. 
% In practice, this means that we construct $B_S \subset \bigcup_{i=1}^{n} H^i(B)$ where $n$ is a hyper parameter controls the depth of rule application. 
Figure~\ref{fig:NLMArch} illustrates this concept, where the knowledge base $B_S$ is constructed by incorporating NLM layers of varying depths. 

\subsubsection{Task Specific Module.}  With the shared knowledge base $B_S$, the task specific module $\widetilde{\cF}_i$ for $i \in \{1, ..., t\}$ is also a NLM layer to take as input all predicates from the knowledge base $B_S$ and compose the target predicates for each corresponding task $\cP_{i}^\star$. Figure~\ref{fig:NLMArch} illustrates this concept, showing how a NLM layer utilizes predicates from the knowledge base to compose the target predicates for each task. 

\subsubsection{Training Protocol.} We facilitate the transfer of the knowledge base across tasks while ensuring that task-specific modules remain distinct and independent from one another.
This means that \emph{the knowledge base is reused from task to task}, while task-specific modules are initialized randomly and trained from scratch. 

\section{Experiments} \label{Experiments}
We note that the transfer of logic rules is beneficial to both supervised learning setting (i.e., ILP) and reinforcement learning setting (i.e., Relational RL). To this end, we present experimental results for both settings to comprehensively demonstrate the value of L2ILP. For supervised learning, we build on insights from the experiments conducted by \cite{dong2019neurallogicmachines}, \cite{zimmer2023differentiablelogicmachines}, \cite{glanois2021neurosymbolichierarchicalruleinduction}, and \cite{li2024llmsrelationalreasoningfar}, proposing task sequences for ILP across three domains: {\bf arithmetic}, {\bf tree}, and {\bf graph}. For reinforcement learning, PDDLGym ~\citep{silver2020pddlgymgymenvironmentspddl} serves as an off-the-shelf tool in which the state and action spaces are represented using logical predicates. We therefore select {\bf BlocksWorld} from PDDLGym as the testbed, as it is a commonly used benchmark for complex logic reasoning tasks~\citep{dvzeroski2001relational, glanois2021neurosymbolichierarchicalruleinduction, valmeekam2023planningabilitieslargelanguage}. \textbf{\textit{We provide the code we used for experiments in the supplementary material.}}
\subsection{Forward Transfer of Logic Rules}

\subsubsection{ILP Experiments.} The key question regarding the proposed approach is whether the transfer of logical rules is genuinely beneficial. We address this by comparing the learning curves of lifelong learning with those of models trained on tasks individually, while keeping model architectures the same. We provide a detailed description of the sequences of target predicates for each domain in the appendix, and present here the designs for two specific domains: Graph and Tree.

In the Graph domain, $IsConnected(X\text{-}Node, Y\text{-}Node)$ and $IsRed(X\text{-}Node)$, fully describe all possible graphs. The first predicate defines the connectivity between nodes, while the second specifies node properties—in this case, the color of the nodes. The four target predicates are learned sequentially in the following order:
\begin{itemize} 
    \item $AdjacentToRed(X\text{-Node})$,
    \item $ExactConnectivity2(X\text{-Node}, Y\text{-Node})$,
    \item $ExactConnectivity2Red(X\text{-Node})$,
    \item $ExactConnectivity2MultipleRed(X\text{-Node})$.
\end{itemize}
The first target predicate determines whether a given node has at least one neighboring node that is red. The second predicate identifies whether the shortest path between two nodes consists of exactly two edges. Building upon this, $ExactConnectivity2Red(X\text{-Node})$ refines the notion of connectivity by verifying whether a node at an exact distance of two from the query node is red. Finally, $ExactConnectivity2MultipleRed(X\text{-Node})$ extends this concept by determining whether there exist multiple such red nodes at the specified distance. We design the tasks in a way that ensures relevance and allows them to share common structures when learning logical rules.

In the Tree domain, $IsParent(X\text{-}Node, Y\text{-}Node)$ is sufficient to specify the structure of any tree. We further define four target predicates to be learned sequentially, following the order below:
\begin{itemize} 
    \item $IsRoot(X\text{-}Node)$
    \item $HasOddEdges(X\text{-}Node, Y\text{-}Node)$
    \item $HasEvenEdges(X\text{-}Node, Y\text{-}Node)$
    \item $IsAncestor(X\text{-}Node, Y\text{-}Node)$
\end{itemize}
The first predicate identifies the root node of the tree and serves as the foundational predicate. The second and the third predicates determine whether two nodes in the tree are connected by an odd or even number of edges, respectively. Finally, $IsAncestor(X\text{-}Node, Y\text{-}Node)$ checks whether one node is an ancestor of another, based on their relative depths in the tree.
\begin{figure}[ht]
    \centering
    
    % Top image (sub-figure)
    \begin{subfigure}{\columnwidth}
        \centering
        \includegraphics[width=1.0\linewidth]{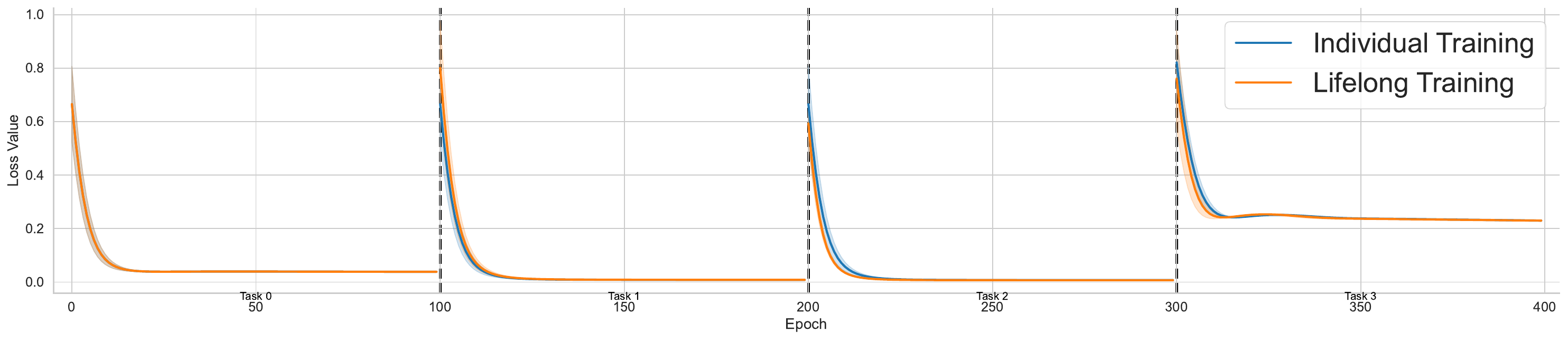} % replace with your image
        \caption{Training dynamics for Arithmetic}
        \label{fig:top}
    \end{subfigure}

    \vspace{1em} % vertical space between the two images
    
    % Bottom image (sub-figure)
    \begin{subfigure}{\columnwidth}
        \centering
        \includegraphics[width=1.0\linewidth]{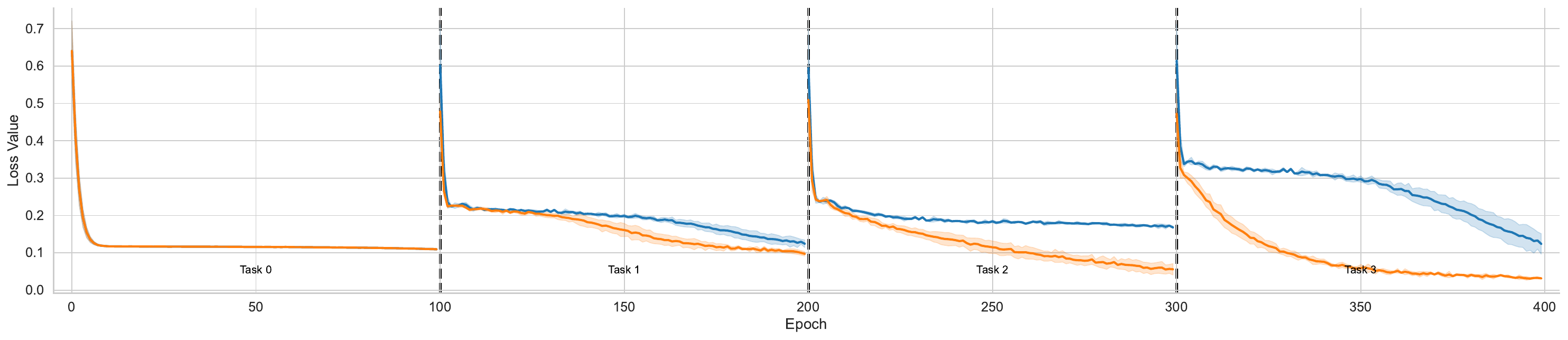} % replace with your image
        \caption{Training dynamics for Tree}
        % \label{fig:bottom}
    \end{subfigure}

    \begin{subfigure}{\columnwidth}
        \centering
        \includegraphics[width=1.0\linewidth]{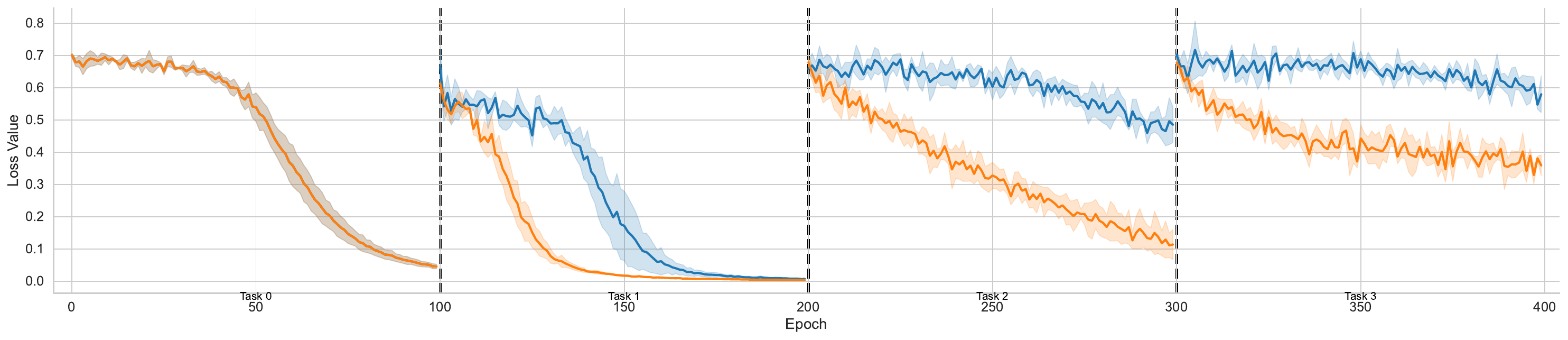} % replace with your image
        \caption{Training dynamics for Graph}
        % \label{fig:bottom}
    \end{subfigure}
    
    \caption{Epoch training dynamics for individual learning and lifelong learning}
    \label{fig:two_images_vertical}
\end{figure}

\begin{figure*}[htbp]
    \centering
    \includegraphics[width=1.0\textwidth]{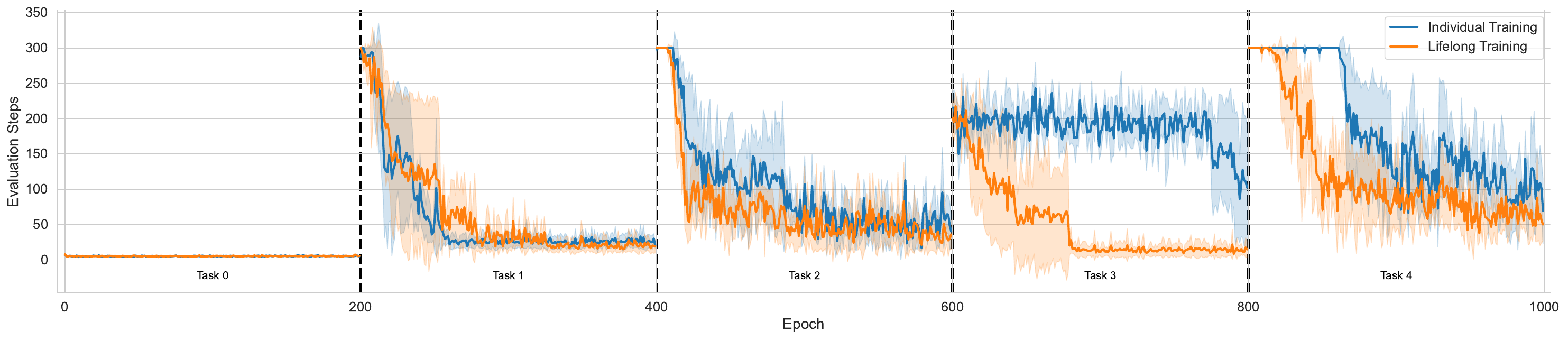}
    \caption{Evaluation steps for BlocksWorld tasks}
    \label{fig:RLLifeLong}
\end{figure*}

Figure~\ref{fig:two_images_vertical} illustrates the training dynamics for each task across all domains. For each task, we conducted experiments with four different seeds and plotted the mean, along with the standard deviation for each data point. From top to bottom, the domains are ordered as arithmetic, tree, and graph, while from left to right, the plots are arranged sequentially for tasks 0 to 3. The tasks in the arithmetic domain are relatively simple, allowing both individual and lifelong learning to converge to optimal performance within a short period. However, we observe that the loss curves in lifelong learning decrease more rapidly, as indicated by a small but noticeable gap. In contrast, for the tree and graph domains, the gaps become more pronounced, with no overlap observed between the paired curves. Notably, for the curves corresponding to the first task across all domains, lifelong learning completely overlaps with individual learning. This is expected, as the first task in lifelong learning does not incorporate any previously acquired knowledge, making it equivalent to its individual learning counterpart.

This result shows that by leveraging the knowledge base acquired in earlier tasks, we could achieve higher learning efficiency on subsequence tasks, demonstrating the \textbf{\textit{forward transfer}} effect that is essential in lifelong learning setting.

\subsubsection{Relational RL Experiments.} For the BlocksWorld task from PDDLGym, Table~\ref{tab:State_Action_Spaces} summarizes the predicates used to describe the state and action spaces. Since the number of ground predicates depends on the number of objects in a task, the total number of possible actions can be as high as $2x^2+2x$, where 
$x$ represents the number of blocks. However, not all of these actions are valid in every state. Additionally, the number of possible states grows factorially with the number of blocks in the environment, making the task increasingly complex as more blocks are added. We note that previous works~\citep{jiang2019neurallogicreinforcementlearning, zimmer2023differentiablelogicmachines, valmeekam2023planningabilitieslargelanguage} typically use \textbf{5} blocks, whereas we extend our experiments to \textbf{6} blocks. Furthermore, We set the tasks to follow a sparse reward scheme, where a penalty of  -0.1 is given for every step taken, while a reward of 100 is granted upon achieving the desired block configuration. To define the sequence of tasks for BlocksWorld, we assign different desired configurations to each task, progressively increasing the difficulty by making the target configurations more challenging to reach. Refer to Appendix for a detailed discussion.

\begin{table}[ht]
  \centering
  \resizebox{0.5\textwidth}{!}{%
    \begin{tabular}{|l|l|l|}
      \hline
      \multicolumn{3}{|c|}{\textbf{State Space}} \\
      \hline
      \textbf{Predicates} & \textbf{Arity} & \textbf{Description} \\
      \hline
      $On(X\text{-}Block, Y\text{-}Block)$     & 2 & Block $X$ is on Block $Y$ \\
      \hline
      $OnTable(X\text{-}Block)$               & 1 & Block $X$ is on the table \\
      \hline
      $Clear(X\text{-}Block)$                 & 1 & Block $X$ could be picked up \\
      \hline
      $HandEmpty()$                           & 0 & Hand is empty \\
      \hline
      $HandFull()$                            & 0 & Hand is full \\
      \hline
      $Holding(X\text{-}Block)$               & 1 & Hand is holding Block $X$ \\
      \hline
      % \multicolumn{3}{|c|}{} \\[-2.5mm] % visual vertical gap with borders
      \hline % ← this is the missing horizontal line above Action Space
      \multicolumn{3}{|c|}{\textbf{Action Space}} \\
      \hline
      \textbf{Predicates} & \textbf{Arity} & \textbf{Description} \\
      \hline
      $PickUp(X\text{-}Block)$                & 1 & Pick up Block $X$ from the table \\
      \hline
      $PutDown(X\text{-}Block)$               & 1 & Put down Block $X$ onto the table \\
      \hline
      $Stack(X\text{-}Block, Y\text{-}Block)$ & 2 & Stack Block $X$ onto Block $Y$ \\
      \hline
      $Unstack(X\text{-}Block, Y\text{-}Block)$ & 2 & Unstack Block $X$ from Block $Y$ \\
      \hline
    \end{tabular}
  }
  \caption{State and action spaces for BlocksWorld from PDDLGym}
  \label{tab:State_Action_Spaces}
\end{table}

Large state and action spaces pose significant challenges for exploration to RL agents, particularly when rewards are sparse and only provided upon goal completion. As a result, our experiments show that training tasks individually fails to yield sufficient exploration to positive rewards, let alone facilitate the learning of any meaningful policies. To address this issue, we adopt an \textbf{offline reinforcement learning} approach, where a replay buffer is collected in advance to ensure that both lifelong learning and individual training have access to data of the same quality. In BlocksWorld tasks, we use a planner to generate optimal actions at each step while incorporating random exploration to ensure adequate coverage of the state and action spaces. Specifically, we set the exploration rate to \textbf{0.8} and collect \textbf{50,000} transitions for each task. 

Figure~\ref{fig:RLLifeLong} illustrates the evaluation steps for each task, as the number of steps directly reflects the quality of the learned policy. We ran each experiment using four different seeds and evaluated the policy periodically. For tasks 0–2, lifelong training does not show superior performance compared to individual training, as these tasks are relatively simple. However, for the more challenging tasks 3 and 4, clear gaps emerge: lifelong training converges much faster, while individual training occasionally fails to converge to the optimal policy. This result highlights that \textbf{\textit{forward transfer}} is not only observed in the relatively straightforward supervised learning setting but also extends to the more complex reinforcement learning paradigm, where effective knowledge transfer accelerates policy learning in later tasks.

\subsection{Forgetting of Logic Rules}
\subsubsection{Forgetting Experiments.} 
\begin{figure}[h]
    \centering
    \includegraphics[width=0.45\textwidth]{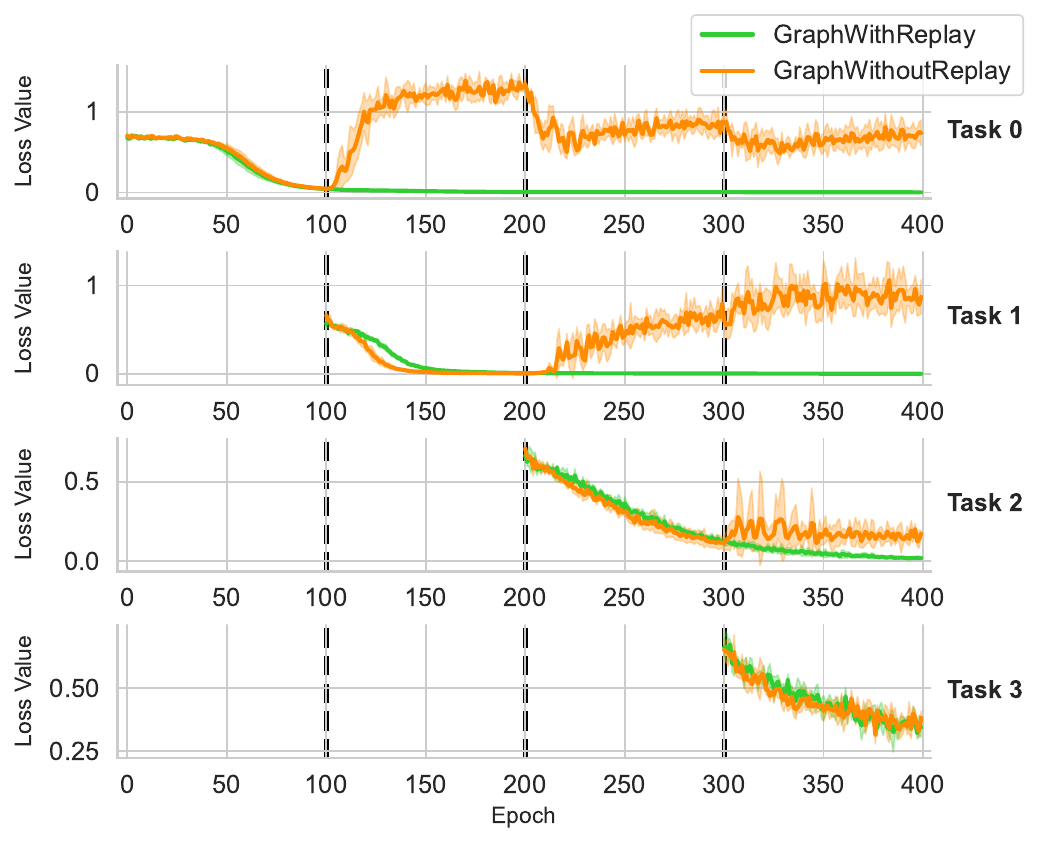}
    \caption{Training dynamics of Graph for each task}
    \label{forgettingGraph}
\end{figure}

Another fundamental challenge in lifelong learning is assessing the extent to which a model forgets previously acquired knowledge (i.e. catastrophic forgetting) and identifying effective strategies to mitigate this issue. In our case, the knowledge base is continuously updated, which may cause task-specific modules for earlier tasks to fail, as they rely on outdated knowledge representations. To investigate this phenomenon, we track the loss values throughout the entire training process and visualize the corresponding curves for each task. Specifically, in the supervised learning setting, we adhere to the training protocol described in the previous section while recording the tested loss values for each task at every epoch whenever feasible. Figure~\ref{forgettingGraph} and Figure~\ref{forgettingTree} illustrate the results for the graph and tree domains, respectively, with each plot representing tasks 0 through 3. For results on arithemetic, please refer to appendix for a detailed description.

In particular, the orange curves illustrate the loss values as described above. As expected, the loss values for earlier tasks increase as training progresses on later tasks, indicating the occurrence of catastrophic forgetting in the model.

\subsubsection{Replay Experiments.} A straightforward approach to addressing catastrophic forgetting is to replay experience from earlier tasks while training on later ones. We adopt this strategy by replaying data from all previously learned tasks when training on a new task. This approach is particularly beneficial in our case, as we aim to build a shared knowledge base that can be effectively leveraged across multiple tasks. To ensure a fair comparison, we align the curves by recording each data point in terms of the training epoch of the current task. This alignment allows us to directly compare the performance of models trained with and without experience replay, providing deeper insights into its effectiveness in mitigating forgetting and preserving previously acquired knowledge.

\begin{figure}[h]
    \centering
    \includegraphics[width=0.45\textwidth]{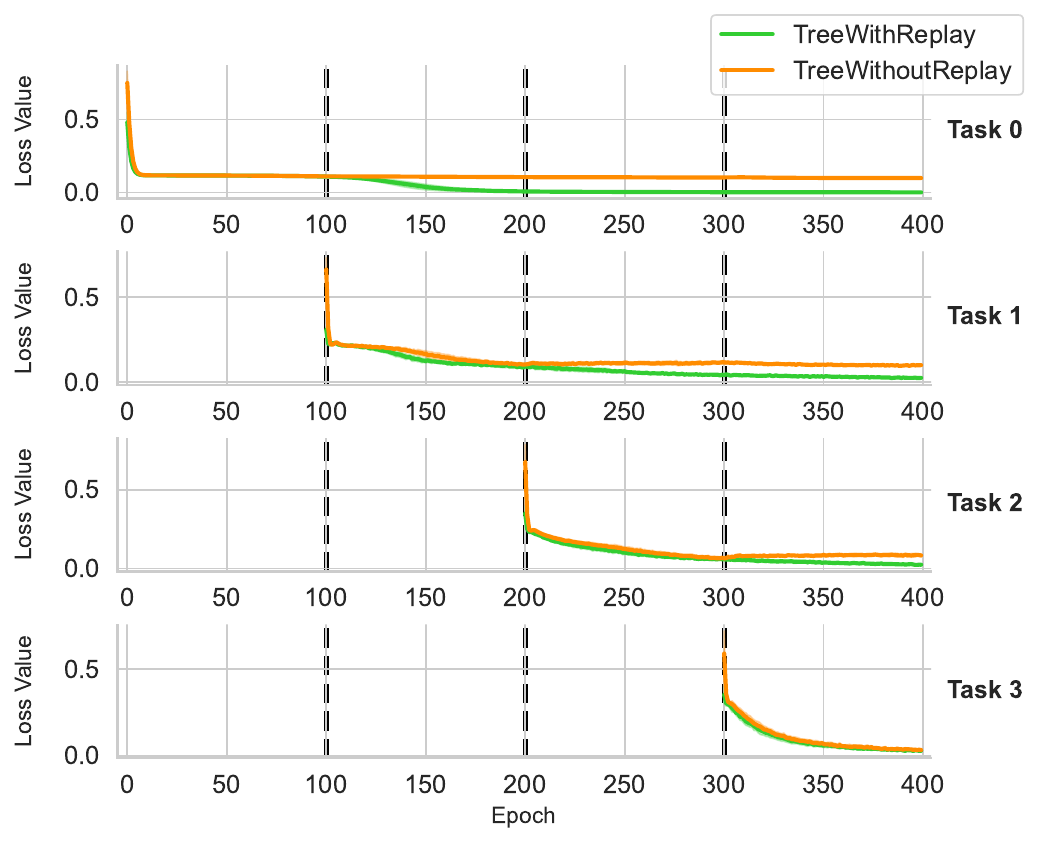}
    \caption{Training dynamics of Tree for each task}
    \label{forgettingTree}
\end{figure}

The green curves in Figure~\ref{forgettingGraph} and Figure~\ref{forgettingTree} illustrate the loss values when experience replay is applied. As the results indicate, during the training phase for each individual task, replaying experience does not significantly interfere with the learning of the current task. The loss curves for training without replay closely align with those for training with replay, indicating that incorporating past experiences does not disrupt the optimization of the current task. More importantly, experience replay effectively mitigates catastrophic forgetting, as evidenced by the green curves maintaining low loss values throughout the entire training process. This suggests that the model successfully constructs 
retains previously acquired knowledge, leading to the formation of a robust and reusable knowledge base across all tasks.

Surprisingly, we also observe a \textbf{\textit{backward transfer}} effect in our experiments. The first plot in Figure~\ref{forgettingTree} corresponds to the loss curve for task 0 in the tree domain. Notably, the loss curve plateaus before training on task 1. However, once training on task 1 begins, the loss for task 0 further decreases to zero, indicating that learning task 1 enhances the model’s performance on task 0. This suggests the potential of L2ILP that \textbf{\emph{training on the current task may further improves the performance on previous tasks, as a better knowledge base is acquired during the sequential training}}.

\section{Conclusion} \label{Conclusion}
Neuro-Symbolic Artificial Intelligence introduces a new research paradigm by integrating neural networks with symbolic methods, which were traditionally studied as separate approaches. This integration opens new research opportunities by enabling the formulation of novel problems and providing solutions to challenges that were previously difficult to address using either approach alone. In this work, we take a step toward studying the lifelong learning problem in this domain and demonstrate that by leveraging the compositionality and transferability of logic rules, it becomes straightforward to construct models that achieve higher learning efficiency on later tasks while preserving performance on earlier ones. However, this problem remains far from solved. As discussed, a key challenge is how to efficiently construct logic rules that are meaningful for tasks within a given domain and how to systematically generate new rules from an evolving knowledge base. Our work represents a small step in this direction, demonstrating its feasibility but leaving many open opportunities for future research. We hope this study inspires further research into more effective methods for representing and constructing logic rules—particularly those that support the dynamic addition and removal of rules from a knowledge base—ultimately enabling more efficient and scalable neuro-symbolic lifelong learning systems.

\newpage
\bibliography{Arxiv}

\clearpage
\onecolumn

% \input{ReproducibilityChecklist
\appendix
\vspace{1em}
\section{Design for the Supervised Learning Tasks}\label{appendix:taskdesign}
\subsection{Arithmetic}
\vspace{1em}
\hspace{2em}\textbf{Input Predicates:}
\begin{equation*}
\begin{aligned}
&\textbf{$Zero(X\text{-}Number)$}      &\quad &\text{True if number $X$ is zero} \\
&\textbf{$Succ(X\text{-}Number, Y\text{-}Number)$} &\quad &\text{True if $Y$ is the successor of $X$}
\end{aligned}
\end{equation*}
\hspace{2em}\textbf{Target Predicates:}
\begin{equation*}
\begin{aligned}
&\textnormal{1.}\textbf{$Plus(X\text{-}Number, Y\text{-}Number, Z\text{-}Number)$}      &\quad &\text{True if $X + Y = Z$} \\
&\textnormal{2.}\textbf{$Times(X\text{-}Number, Y\text{-}Number, Z\text{-}Number)$}     &\quad &\text{True if $X \times Y = Z$} \\
&\textnormal{3.}\textbf{$Division(X\text{-}Number, Y\text{-}Number, Z\text{-}Number)$}  &\quad &\text{True if $X \div Y = Z$ (integer division)} \\
&\textnormal{4.}\textbf{$NoRemainder(X\text{-}Number, Y\text{-}Number)$}                &\quad &\text{True if $X$ is divisible by $Y$ with no remainder}
\end{aligned}
\end{equation*}
\subsection{Tree}
\vspace{1em}
\hspace{2em}\textbf{Input Predicates:}
\begin{equation*}
\begin{aligned}
&\textbf{$IsParent(X\text{-}Node, Y\text{-}Node)$} &\quad &\text{True if $X$ is the parent of $Y$}
\end{aligned}
\end{equation*}
\hspace{2em}\textbf{Target Predicates:}
\begin{equation*}
\begin{aligned}
&\textnormal{1.}\textbf{$IsRoot(X\text{-}Node)$}                      &\quad &\text{True if $X$ is the root node} \\
&\textnormal{2.}\textbf{$HasOddEdges(X\text{-}Node, Y\text{-}Node)$}  &\quad &\text{True if the path from $X$ to $Y$ has an odd number of edges} \\
&\textnormal{3.}\textbf{$HasEvenEdges(X\text{-}Node, Y\text{-}Node)$} &\quad &\text{True if the path from $X$ to $Y$ has an even number of edges} \\
&\textnormal{4.}\textbf{$IsAncestor(X\text{-}Node, Y\text{-}Node)$}   &\quad &\text{True if $X$ is an ancestor of $Y$}
\end{aligned}
\end{equation*}

\subsection{Graph}
\vspace{1em}
\hspace{2em}\textbf{Input Predicates:}
\begin{equation*}
\begin{aligned}
&\textbf{$IsConnected(X\text{-}Node, Y\text{-}Node)$} &\quad &\text{True if there is an edge between $X$ and $Y$} \\
&\textbf{$IsRed(X\text{-}Node)$}                      &\quad &\text{True if node $X$ is red}
\end{aligned}
\end{equation*}
\hspace{2em}\textbf{Target Predicates:}
\begin{equation*}
\begin{aligned}
&\textnormal{1.}\textbf{$AdjacentToRed(X\text{-}Node)$}                      &\quad &\text{True if node $X$ is connected to a red node} \\
&\textnormal{2.}\textbf{$ExactConnectivity2(X\text{-}Node, Y\text{-}Node)$}  &\quad &\text{True if there is a path of exactly 2 edges from $X$ to $Y$} \\
&\textnormal{3.}\textbf{$ExactConnectivity2Red(X\text{-}Node)$}             &\quad &\text{True if $X$ is connected via exactly 2 edges to a red node} \\
&\textnormal{4.}\textbf{$ExactConnectivity2MultipleRed(X\text{-}Node)$}     &\quad &\text{True if $X$ is connected via exactly 2 edges to multiple red nodes}
\end{aligned}
\end{equation*}

\vspace{1em}
\section{Design for the reinforcement learning tasks}\label{appendix:rltaskdesign}
\vspace{1em}
The sequence of tasks in reinforcement learning varies based on their target configurations. For each task, we define specific configurations without considering the order of the blocks and sample a target configuration at the beginning of each episode. For example, if a task’s target configuration requires three blocks to be on the table, two blocks to be stacked on others, and one block to be freely placed anywhere, we first generate possible configurations that meet this requirement and then sample a target configuration accordingly. This task design provides the flexibility to configure the target configuration space, allowing us to adjust the difficulty of the tasks as needed.\\
\\
Thus, we define the five tasks in the following order:
\begin{itemize}
    \item 5 blocks are on table, 1 block is free
    \item 3 blocks on table, 2 blocks are stacked, 1 block is free
    \item 2 blocks on table, 1 block is stacked, 1 block is free
    \item 4 blocks on table, 1 block is stacked, 1 block is free
    \item 1 block is on table, 4 blocks are stacked, 1 block is free
\end{itemize}
The order of the tasks are determined by the performance of the model trained on those tasks individually.
\vspace{1em}
\section{curves for forgetting experiments on arithmetic}\label{appendix:forgettingresults}
\vspace{1em}
\begin{figure}[h]
    \centering
    \includegraphics[width=0.6\textwidth]{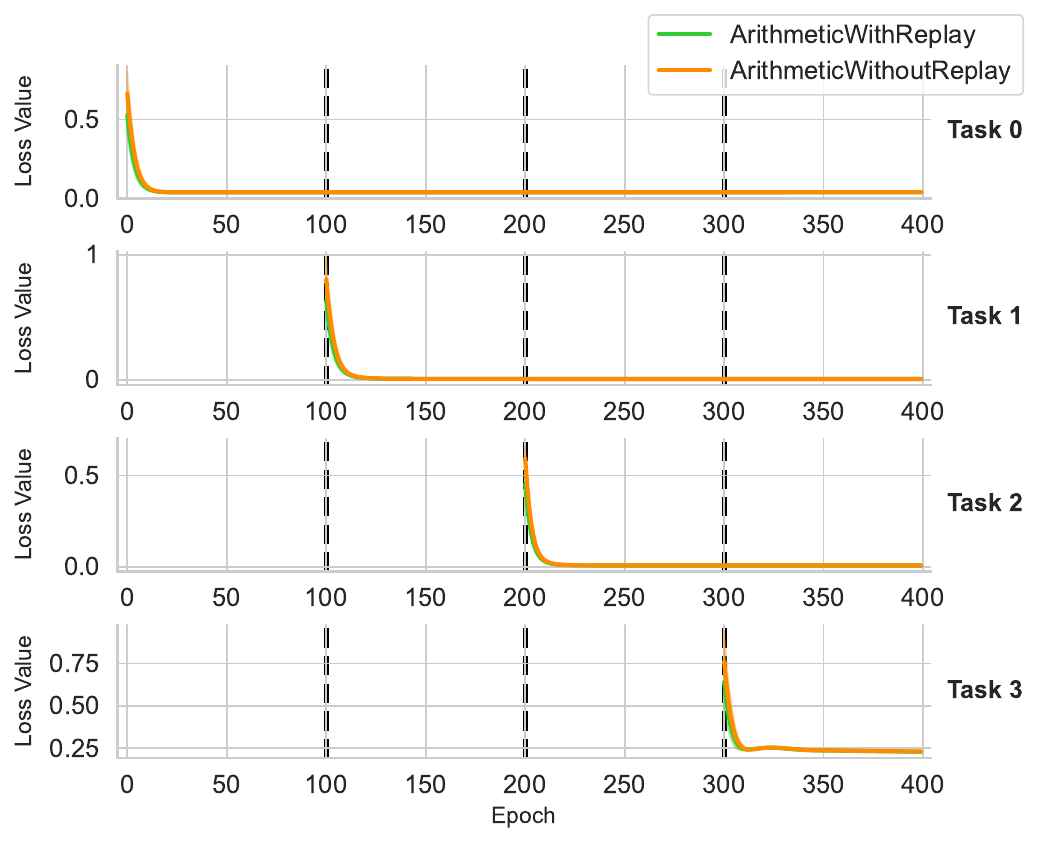}
    \caption{Loss Curves of Arithmetic for each task}
    \label{forgettingArithmetic}
\end{figure}

\vspace{1em}
\section{Task Hyper parameter setting}
The choice of task generation hyper parameters impacts the experimental results. Hyper parameters that make the tasks too easy may lead to trivial performance differences, while excessively difficult tasks can hinder learning altogether. We thus report the choice of task generation hyper parameters we selected in the experiments.
\vspace{1em}
\subsection{Arithmetic}
\begin{itemize}
    \item \textbf{Range of Numbers:} 0 to 79
\end{itemize}
\vspace{1em}
\subsection{Tree}
\begin{itemize}
    \item \textbf{Number of Nodes:} 40
    \item \textbf{Maximum Children per Node:} 3
    \item \textbf{Minimum Children per Node:} 2
\end{itemize}
\vspace{1em}
\subsection{Graph}
\begin{itemize}
    \item \textbf{Number of Nodes:} 30
    \item \textbf{Maximum Edge Generation Probability:} 0.1
    \item \textbf{Minimum Edge Generation Probability:} 0.01
\end{itemize}
\vspace{1em}
\section{Model Hyper parameters}
\vspace{1em}
\subsection{ILP}
\begin{itemize}
    \item \textbf{Knowledge Base Settings:}
    \begin{itemize}
        \item $[[8, 8, 8, 8]]$ – 8 predicates of arity 0 through 3, 1 layer
        \item $[[8, 8, 8, 8], [8, 8, 8, 8]]$ – 8 predicates of arity 0 through 3, 2 layers
        \item $[[8, 8, 8, 8], [8, 8, 8, 8], [8, 8, 8, 8]]$ – 8 predicates of arity 0 through 3, 3 layers
        \item $[[8, 8, 8, 8], [8, 8, 8, 8], [8, 8, 8, 8], [8, 8, 8, 8]]$ – 8 predicates of arity 0 through 3, 4 layers
    \end{itemize}
    
    \item \textbf{Task-Specific Module Setting:}
    \begin{itemize}
        \item $[[8, 8, 8, 8]]$ – 8 predicates for arity 0 to 3 for composing the target predicate
    \end{itemize}
    
    \item \textbf{Learning Rate:} 0.001
\end{itemize}
\vspace{1em}
\subsection{RL}
\begin{itemize}
    \item \textbf{Knowledge Base Settings:}
    \begin{itemize}
        \item $[[4, 4, 4, 4]]$ – 4 predicates of arity 0 through 3, 1 layer
        \item $[[4, 4, 4, 4], [4, 4, 4, 4]]$ – 4 predicates of arity 0 through 3, 2 layers
        \item $[[4, 4, 4, 4], [4, 4, 4, 4], [4, 4, 4, 4]]$ – 4 predicates of arity 0 through 3, 3 layers
        \item $[[4, 4, 4, 4], [4, 4, 4, 4], [4, 4, 4, 4], [4, 4, 4, 4]]$ – 4 predicates of arity 0 through 3, 4 layers
    \end{itemize}
    
    \item \textbf{Task-Specific Module Setting:}
    \begin{itemize}
        \item $[[8, 8, 8, 8]]$ – 8 predicates of arity 0 to 3 for composing the action predicates
    \end{itemize}
    
    \item \textbf{Actor Learning Rate:} 0.003
    \item \textbf{Critic Learning Rate:} 0.003
    \item \textbf{Target Network Copy Rate:} 0.05
    \item \textbf{Actor Update Delay:} 0.05
    \item \textbf{Temperature for Generating Action Distribution:} 1.5
    \item \textbf{Activation for generating Action Logits:} tanh
    \item \textbf{Collected Buffer Size:} 50,000
    \item \textbf{Random Exploration Rate when Collecting Transitions:} 0.8
\end{itemize}
\vspace{1em}
\section{Background for Logic Programming and ILP}
This section includes some background information about \textit{logic programming}, followed by its learning counterpart, \textit{inductive logic programming}. \textbf{\emph{Note that the material is primarily adapted from existing sources and can be readily obtained from other resources such as~\cite{cropper2022inductivelogicprogramming30}}}.
\vspace{1em}
\subsection{Logic Programming}
Logic programming is a programming paradigm that execute based on formal logic sentences, rather than commands or functions typical of conventional programming. Knowledge about a domain is typically represented using logic facts and rules in first-order logic, with computation performed by iteratively applying these rules to deduce new facts. Here we include the definitions and concepts that are commonly used in logic programming.

\textbf{Predicate} A predicate $p$ characterizes a property of an object or describes a relationship among multiple objects. It functions as a Boolean operator, returning either $True$ or $False$ based on its inputs. Consequently, a predicate can be represented as $p(t)$ in the case of a single-variable predicate or $p(t_1, t_2, ...)$ for predicates involving multiple variables. 

\textbf{Atoms} An atom is formed by expressing a predicate along with its terms, which can be either constants or variables. It becomes grounded when all its inputs are replaced by specific constants (objects) within a domain, thereby establishing a definite relationship among these objects, and can be evaluated as $True$ or $False$. As the fundamental unit in logical expressions, it is aptly termed "atom".

\textbf{Literal} A literal is an atom or its negation, representing an atomic formula that contains no other logical connectives such as $\land$, $\lor$. 

\textbf{Clauses} A clause is constructed using a finite set of literals and connectives. The most commonly used form of clause is the \textbf{definite clause}, which represents a disjunction of literals with exactly one positive literal included. It is expressed as
\begin{equation*}
    p_1 \lor \neg p_2 \lor \neg p_3 ... \lor \neg p_n
\end{equation*}
The truth value of the expression above is equivalent to that of an implication, allowing it to be rewritten in the form of an implication as follows:
\begin{equation*}
    p_1 \leftarrow p_2 \land p_3 ... \land p_n
\end{equation*}
The part to the left of the implication arrow is typically referred to as the \textbf{head} of the clause, while the part to the right is known as the \textbf{body}. A clause is read from right to left as the head is entailed by the conjunction of the body.
A clause is \textbf{grounded} when all its literals are instantiated with constants from a domain. A ground clause without a body constitutes a \textbf{fact} about the domain, defined by the specific predicate and associated objects. A clause that includes variables is often treated as a rule about a domain.

Logic programming involves defining a set of clauses $R$ including both rules and facts. The consequences of the set $R$ is computed by repeatedly applying the rules in $R$ until no more facts could be derived, where a convergence is considered to have been reached. We take the example from \cite{evans2018learningexplanatoryrulesnoisy} to give an illustration on logic programming. Consider the program R as
\begin{equation*}
\begin{aligned}
    edge(a, b) &\quad edge(b, c) \quad edge(c, a)\\
    connected(X, Y) &\leftarrow edge(X, Y)\\
    connected(X, Z) &\leftarrow edge(X, Y) \land connected(Y, Z)
\end{aligned}
\end{equation*}
where $\{a, b, c, d\}$ represents the set of objects considered in the domain, while $\{X, Y, Z\}$ are variables in each predicate rule. Thus, $\{edge(a. b), edge(b, c), edge(c, d\}$ are considered ground facts while the other clauses are considered logic rules that could be applied to derived new facts about a domain.

The computation of consequences of the program could be summarized as
\begin{equation*}
\begin{aligned}
    C_{R, 1} &= \{edge(a,b), edge(b,c), edge(c,a)\}\\
    C_{R, 2} &= C_{R,1} \cup \{connected(a, b), connected(b, c), connected(c, a)\}\\
    C_{R, 3} &= C_{R, 2} \cup \{connected(a, c), connected(b, a), connected(c, b)\}\\
    C_{R, 4} &= C_{R, 3} \cup \{connected(a, a), connected(b, b), connected(c, c)\}\\
    C_{R, 5} &= C_{R, 4} = con(R)
\end{aligned}
\end{equation*}
Thus, from the single-direction edge predicate and the rules defining the connection between two nodes, we deduce that all nodes are connected, including self-connections. $C_{R, 5}$ constitutes the final conclusions of the program, as no additional facts are added beyond $C_{R, 4}$.

\vspace{1em}
\subsection{Inductive Logic Programming}
Inductive logic programming(ILP) aims for the opposite goal of logic programming: given a set of facts, it seeks to derive the rules that best explain those facts. We include several more definitions here to better explain the topic.

\textbf{Background and Target Predicates} We categorize the predicates involved in a problem into two sets: background predicates, which provide the foundational facts for logical deductions, and target predicates, which represent the conclusions derived from these deductions.

\textbf{Extensional and Intentional Predicates} In the context of clauses, extensional predicates never appear in the heads of clauses, whereas intentional predicates can. Background predicates, often the starting points for logical deduction, are typically extensional. In contrast, target predicates, which result from logical deductions, are usually intentional. The concept of intentional predicates is crucial, as ILP often requires inventing intermediary intentional predicates to facilitate program solving.

An inductive logic programming problem is defined by a tuple $\{B, E^+, E^-\}$ of ground atoms, where$B$ specifies the background knowledge, which consists of ground atoms formed from the set of background predicates. $E^+$ and $E^-$ denote the positive and negative targets, respectively, which are ground atoms formed from the set of target predicates.

The solution to an ILP involves finding a set of lifted rules $U$ such that,
\begin{align*}
    \forall e \in E^+, B \cup U |\hspace{-.5em}= e \\
    \forall e \in E^-, B \cup U \cancel{|\hspace{-.5em}=} e
\end{align*}
where $|\hspace{-.7em}=$ denotes logic entailment. More intuitively, this means that ground atoms from $E^+$ are included in the deduction consequences, while ground atoms from $E^-$ are excluded.

We take an example from~\cite{evans2018learningexplanatoryrulesnoisy} to illustrate the idea. Suppose a ILP problem represented as
\begin{equation*}
\begin{aligned}
    B &= \{zero(0), succ(0, 1), succ(1, 2), ...\}\\
    E^+ &= \{even(0), even(2), even(4), ...\}\\
    E^- &= \{even(1), even(3), even(5), ...\}\\
\end{aligned}
\end{equation*}
where the predicates $zero()$ and $succ()$ serve as background predicates that facilitate the definition of the knowledge base, while the predicate $even()$ defines the learning target as the target predicate. A possible solution $R$ for the ILP could be
\begin{equation*}
\begin{aligned}
    even(X) &\leftarrow zero(X)\\
    even(X) &\leftarrow even(Y) \land succ2(Y, X)\\
    succ2(X, Y) &\leftarrow succ(X, Z) \land succ(Z, Y)\\
\end{aligned}
\end{equation*}
An obvious deduction process would be
\begin{equation*}
\begin{aligned}
    C_{R, 1} &= B\\
    C_{R, 2} &= C_{R,1} \cup \{even(0), succ2(0, 2), succ2(2, 4), succ2(4, 6), ...\}\\
    C_{R, 3} &= C_{R, 2} \cup \{even(2), even(4), even(6), ...\}\\
    C_{R, 4} &= C_{R, 2} = con(R)
\end{aligned}
\end{equation*}
Clearly, it is satisfied that $\forall e \in E^+, e \in con(R)$ and $\forall e \in E^-, e \notin con(R)$, thus the solution is accepted. The above example is not totally trivial, as it illustrates the concept of \textbf{predicate invention}: $succ2()$ is synthesised to facilitate logic deduction. To connect the example with the concepts defined in the beginning of the section, predicates $zero()$ and $succ()$ are extensional predicates as they never have to be deduced from the rules while $succ2()$ and $even()$ are intentional as they appear in the deduction heads of the rules.

\end{document}